\documentclass[10pt]{article} 
\usepackage{tmlr}

\usepackage{amsmath,amsfonts,bm}

\def\eqref#1{equation~\ref{#1}}

\def\1{\bm{1}}

\DeclareMathAlphabet{\mathsfit}{\encodingdefault}{\sfdefault}{m}{sl}
\SetMathAlphabet{\mathsfit}{bold}{\encodingdefault}{\sfdefault}{bx}{n}

\usepackage[hidelinks]{hyperref}

\usepackage{url}

\usepackage{amssymb}

\usepackage{enumitem}

\usepackage{algorithm}
\usepackage{algpseudocode} 

\usepackage{graphicx}
\usepackage{pgfplots}
\pgfplotsset{compat=1.18}

\title{Source-Free Domain Adaptation by Optimizing Batch-Wise Cosine Similarity}

\author{\name Harsharaj Pathak \email harsharaj.pathak.ete@gmail.com \\
      \addr Machine Learning and Vision Lab\\
      Indian Institute of Technology Hyderabad
      \AND
      \name Vineeth N Balasubramanian \\
      \addr Department of Computer Science and Engineering\\
      Indian Institute of Technology Hyderabad}

\def\openreview{\url{https://openreview.net/forum?id=XXXX}} 

\begin{document}

\maketitle

\begin{abstract}
Source-Free Domain Adaptation (SFDA) is an emerging area of research that aims to adapt
a model trained on a labeled source domain to an unlabeled target domain without accessing
the source data. Most of the successful methods in this area rely on the concept of neighbor-
hood consistency but are prone to errors due to misleading neighborhood information. In
this paper, we explore this approach from the point of view of learning more informative clusters
and mitigating the effect of noisy neighbors using a concept called neighborhood signature,
and demonstrate that adaptation can be achieved using just a single loss term tailored
to optimize the similarity and dissimilarity of predictions of samples in the target domain.
In particular, our proposed method outperforms existing methods in the challenging VisDA
dataset while also yielding competitive results on other benchmark datasets.
\end{abstract}

\section{Introduction}

Deep learning models trained on labeled data are known to perform very well on a wide range of tasks. However, one key assumption for most models is that the test data comes from the same distribution as the training data \citep{theolearn, stleatheo}. These distributions are also referred to as "domains," as data collected from a particular domain tends to follow some specific distribution. When the test domain is different from the training domain, the performance of the models deteriorate, often drastically, due to the distribution shift \citep{coshiftdis, dshiftml}. As a result, there has been a growing body of research in the area of Unsupervised Domain Adaptation (UDA) which seeks to adapt a model trained on a source domain to a target domain using unlabeled data from the target domain, which are generally easier to collect than labeled data \citep{unsuback, geodes, pixlevel, domadver}. However, most UDA methods also require the labeled data from the source domain during the adaptation \citep{frusda, conda, dirty}. This can be impractical in many scenarios where we only have access to the source model and not the source data due to privacy or copyright reasons. To tackle this, a more restrictive setting has been explored, namely, Source-Free Domain Adaptation (SFDA) where we only have access to the trained source model and the unlabeled target data.  

In this paper, we specifically focus on SFDA in the context of image classification. Several methods have thus far been explored for SFDA in image classification \citep{unsfda, danosource, doneeddata, testdisshift, doubleattend, goodfriend, aada}. A defining characteristic of many of these methods is the reliance on neighborhood consistency in the feature space \citep{doneeddata, goodfriend, guideda}. The assumption here is that despite the domain shift, the source model forms clusters based on the semantic similarity of the samples from the target domain in the feature space \citep{goodfriend} and this can be used to encourage similar predictions for samples that are neighbors of one another in the feature space. Existing methods often leverage this through techniques such as estimating pseudo labels by aggregating the predictions of neighbors \citep{doneeddata, guideda} or directly enforcing neighborhood consistency with multiple neighbors \citep{goodfriend, aada, dalensaug} along with additional objectives such as diversity \citep{doneeddata} and contrastive loss \citep{guideda, daoverlook}. 

The key limitation of these methods is that they are prone to learning misleading labels due to noisy neighbors, as there is no guarantee that most neighboring samples of a sample are correctly classified at any point in the training. In this paper, we propose a new method for this task to mitigate the effect of noisy neighbors. Our key idea is to use neighborhood consistency in a more reliable manner. To this end, we propose some key insights that can be used to estimate the pairwise semantic similarity and dissimilarity of samples in a training batch in the target domain, which ultimately leads to the model estimating the classes of these samples. 

Our core approach is to align the predictions of the samples in a batch with respect to cosine similarity, such that the samples that are estimated to be more semantically similar have greater cosine similarity in their predictions and vice versa instead of directly mapping samples into classes. This encourages the model to consider a more holistic view of the semantic information and enables more flexibility in dealing with difficult samples that have overlapping semantic characteristics with multiple classes. In order to estimate the semantic similarity, we introduce a scheme called neighborhood signature, which in our case is simply the mean prediction of the neighborhood samples, but unlike previous methods, we don’t use the neighbors to directly estimate training pseudo labels. Another idea we propose is to encourage the model to learn more informative clusters, i.e., learn semantic differences of samples even when they are estimated to belong to the same class. The goal is to prevent the model from taking shortcuts and assigning samples to classes based on the most common semantic characteristics of the class and thus avoid confidently fitting samples to the wrong class due to noisy neighbors. For example, a model may learn to classify any yellow vehicle as a bus while ignoring the possibility of it being a car or a truck. The model has a better chance of avoiding this if it recognizes the variations within the bus, car, and truck classes. This can also implicitly prevent mode collapse without a separate diversity objective. The third consideration in our method is to consider the confidence with which samples are predicted and use this as inertia while aligning the predictions. For example, if we believe that two objects are semantically similar, and we are confident that one is a cat, and we are less confident that the other is a dog, then it is more likely that both are cats rather than both are dogs. Finally, we also consider the effect of class imbalance. Since different classes may have different frequencies of occurrence while aligning the predictions, the classes with greater frequency can have a dominating effect on the other classes; hence, we counter this by scaling the effect of samples based on their estimated class and the estimated frequency of the classes, even though these estimated classes are not directly used as training labels. Despite all of the above factors, our method still boils down to a single loss term where all the above factors are encoded in a mask that dictates the strength with which the pairwise cosine similarities of the predictions of samples should increase or decrease. Experimental results on benchmark data show the efficacy of our method, notably outperforming existing methods in the challenging VisDA dataset.

In Summary, our contributions are:

\begin{itemize}[label=$\blacksquare$]
\item We propose a novel approach to the SFDA problem that involves aligning cosine similarities of sample predictions based on neighborhood consistency, intra-class diversity, confidence, and class imbalance, all encoded into a single loss term.
\item We conduct experiments on benchmark data and demonstrate the effectiveness of our method.
\end{itemize}

\section{Related Work}
\label{rel_wrk}
\textbf{Domain Adaptation} Early Domain Adaptation methods mostly relied on the idea of aligning the feature distributions of the source and target domains. CORAL\citep{coral} uses moment matching. DANN \citet{domadver} uses adversarial training to fool a domain discriminator while maintaining discriminative-ness of features. CDAN \citet{conda} uses conditioning on discriminative information to assist adversarial feature alignment. DIRT-T \citet{dirty} uses adversarial training with the added constraint that samples belonging to clusters in the target domain belong to the same class. Additionally, \citet{sliceda, stoiclass, maxdiscrep} are based on the idea of employing multiple classifier heads to give correct outputs in source domain and diverse outputs in the target domain and then aligning the domains through adversarial training. More recently, self supervision based methods have gained more traction. These methods often leverage pseudo labels generated by the model in the target domain. One early example is CST \citet{cycleda} which improves pseudo labels through a reverse training strategy based on making the pseudo labels more useful for the source model. 

\textbf{Source-free Domain Adaptation} In Source-free Domain Adaptation, the source training data is unavailable during the adaptation, and so we must rely only on the source model. Most well known methods in this setting rely on self supervision. Early methods such as USFDA \citet{unsfda} and FS \citet{dainherit} leverage synthetic samples. SHOT \citet{doneeddata} learns pseudo labels based on nearest class centroids combined with entropy minimization and diversity maximization. 3C-GAN \citet{danosource} collaborates between a conditional generator and a classifier to gradually adapt the classifier to target styled samples. $A^2$\text{-Net} \citet{doubleattend} employs separation of source similar and source dissimilar samples and applies category wise matching. NRC \citet{goodfriend} is a pseudo label based method that matches predictions with nearest neighbours while prioritizing reciprocal neighbours. AAD \citet{aada} introduces the the idea of maximizing prediction mismatch with background samples. \text{SF(DA)$^2$} \citet{dalensaug} applies data augmentation in the latent feature space based on augmentation graphs and learns high quality clusters. SiLAN \citet{daoverlook} is another latent augmentation based method that uses contrastive learning and guides the augmentation using the feature dispersion statistics of the source model. While these methods employ various techniques to incorporate additional information in the training and improve cluster reliability, none can fully eliminate the problem of noisy neighbours. Our method explores an alternative approach which can move this field forward by improving the utilization of neighbourhood information. 

\textbf{Deep Clustering} Deep Clustering refers to clustering very high dimensional data by first reducing them to a low dimensional feature space using a deep neural network. Earlier methods include DAC \citet{deepadaptclust} which does self supervised clustering of images based on pairwise similarity estimates which are in turn adapted by training a deep neural network. DCCM \citet{deepcomprehend} also uses pairwise similarity while improving the category estimates using pseudo labels along with maximizing local transformation robustness and mutual information. DEC \citet{deepembedanayse} is an auto-encoder based deep clustering method which assigns soft clusters based on cluster centroids and optimizes the KL-divergence between the clusters and an auxiliary target distribution. CC \citet{contraclust} applies contrastive learning at both instance level and cluster level to directly map instances to cluster vectors.

\section{Method}
\label{metd}

For source-free domain adaptation (SFDA), we are given a source-pretrained model and an unlabelled dataset with $N_t$ samples from the target domain as $D_t = \{\bm{x_i}^t\}_{i=0}^{N_t}$. Both the source domain and the target domain have the same $C$ classes. The goal is to adapt the model to the target domain. We denote the model as consisting of two parts: the feature extractor $f$, and the classifier $g$. The output of the feature extractor is denoted as the feature $(\bm{z_i} = f(\bm{x_i}) \in \mathbb{R}^h)$, where h is the dimension of the feature space. The output of the classifier is denoted as $(\bm{p_i} = \delta(g(\bm{z_i})) \in \mathbb{R}^C)$ where $\delta$ is the soft max function. We denote $P \in \mathbb{R}^{bs \times C}$ as the prediction matrix in a mini-batch, where $bs$ is the batch size. In order to do the adaptation, we formulate a loss function based on the cosine similarity of the prediction of each sample and the class encoding of all other samples in the mini batch. The loss function can be represented as:

\begin{equation}
L = \sum_{\substack{0 \leq i < bs \\ 0 \leq j < bs \\ i \neq j}} \bm{p_i} \bm{q_j} M_{ij}
\end{equation}

where $\bm{p_i}$ is the model prediction of the $i$\textsuperscript{th} sample and $\bm{q_j}$ is the class encoding (discussed later) of the $j$\textsuperscript{th} sample and $M_{ij}$ is the mask that determines the magnitude and direction of the rate at which $\bm{p_i}$ and $\bm{q_j}$ should align.

We now describe the class encoding $\bm{q_j}$ and the various components of the mask $M_{ij}$

\textbf{Neighborhood Signature} As mentioned previously, we are leveraging the concept of neighbourhood consistency, i.e. we expect the model initialized as the source model to extract meaningful features in the target domain which causes semantically similar samples to map close to one another in the feature space. However, instead of directly estimating training labels from the potentially noisy neighbourhood, we work with the assumption that semantically similar samples have similar neighbourhood. The intuitive basis for this assumption is that, if the model confuses a tree beside a road with a lamp post beside a road then the same model is likely to confuse a tree beside a house with a lamp post beside a house and hence both tree samples would end up with similar looking lamp post samples in their neighbourhood even if the tree samples are not close to one another because one is influenced by road features and the other is influenced by house features (as the model is not already adept in extracting only relevant features in the target domain). Thus, we can gain meaningful information about the semantics of a sample by observing its neighbours. We refer to this information as neighbourhood signature. The question now is, how to encode the neighbourhood information to achieve this signature ? This can be explored in details in future work, but for the purpose of this paper we use the most obvious encoding, which is taking the mean prediction of the neighbourhood samples.  Thus, for any given feature $\bm{z_i}$, its neighbourhood signature is given by:
\begin{equation}
\bm{s_i} = \frac{\sum_{k \in N(i)} \bm{p_k}}{|N(i)|}
\end{equation}

where $N(i)$ is the set of  neighbours of the $i$\textsuperscript{th} sample and $\bm{p_k}$ is the model prediction for the $k$\textsuperscript{th} sample as stored in the score bank. Note that, taking mean prediction of neighbours is a common approach, but our method differs in the motivation for it and the  way it is used.

\textbf{Confidence-Adaptive Class Encoding} The class encoding for a sample is a tensor that encodes the estimated semantic information in a sample in terms of the probability of it belonging to each of the classes. In this sense it is similar to the pseudo label which is used in domain adaptation methods but in our case we do not try to fit samples to their own class encoding, rather we align the predictions of samples with the class encoding of other samples in a mini batch based on their estimated semantic similarity. One choice for class encoding is to use the neighbourhood signature itself since in our case it of the same dimension as the output probabilities, but this risks reinforcing noisy neighbourhoods, thus leading to misleading semantic estimates. To resolve this, we use a simple solution which is to choose either the current prediction of the sample or the neighbourhood signature, depending on which ever has the lesser prediction entropy. Therefore, the class encoding $\bm{q_j}$ of the $j$\textsuperscript{th} sample is given by:

\begin{equation}
\bm{q_j} = \arg\min_{\bm{x} \in \{\bm{s_j}, \bm{p_j}\}} H(\bm{x})
\end{equation}

\textbf{Neighborhood Similarity} Now, in order to construct the mask $M_{ij}$, we use the hypothesis that the cosine similarity between the prediction of the $i$\textsuperscript{th} sample $\bm{p_i}$ and a class encoding $\bm{q_j}$ in the loss should increase if samples $i$ and $j$ have similar neighbourhood and vice versa. The reasoning, as described before, is that semantically similar samples should have similar neighbourhoods. Therefore, one proposed component of the mask is: 
\begin{equation}
M_{ij1} = 1 - 2 \cdot \bm{s_i} \cdot \bm{s_j}
\end{equation}

As we can observe, for very similar neighbourhood ($\bm{s_i} \cdot \bm{s_j}$ approaching 1), this component approaches -1 which will strongly align $\bm{p_i}$ and $\bm{q_j}$ as we minimize the loss while for very dissimilar neighbourhood ($\bm{s_i} \cdot \bm{s_j}$ approaching 0) this component will approach 1 thus causing $\bm{p_i}$ to move away from $\bm{q_j}$. 

Since we do not yet know how to perfectly encode the neighbourhood information so that our mask in Eq 1. equals the ideal mask, we improve upon the raw masks achieved by our mean based neighbourhood similarity with the following additional terms based on practical observations. These terms are not exhaustive, and the core idea provides the scope to add more terms and engineer the mask for better adaptation. 

\textbf{Intra-Class Diversity} As mentioned earlier, we want to encourage intra-class diversity to encourage the model to learn semantic variations within a class. For this, we add a second component to the mask $M_{ij}$ as:
\begin{equation}
M_{ij2} = \alpha \cdot \bm{s_i} \cdot \bm{s_j}
\end{equation}

Here, $\alpha$ is a decay term that is initialized as 1 and decays to 0 as the training progresses. This term forces samples with similar neighbourhood to be unaligned and the combined effect of this and the previous term leads to a weak repulsive force between samples which are semantically similar and likely belong to the same class with this repulsion being stronger for more dissimilar samples. This term is decayed out into order to prevent it from hindering the formation of compact clusters towards the end of the training.

\textbf{Inertia of Class Encoding} We propose that when a prediction is being aligned with respect to the class encoding of another sample (whether aligning towards or away), the strength of this alignment should depend on the confidence of the class encoding. Consider two samples $i$ and $j$ which are being estimated to be similar based on their neighbourhood signatures. If they have different class encoding, then it is more likely that the one with the more confident class encoding represents the correct label for both hence the prediction for the one with the less confident class encoding should be moved with greater strength towards the class encoding of the other sample. The same is true while moving predictions away from a class encoding if they are estimated to be dissimilar based on their neighbourhood signature. Thus, we scale the mask $M_{ij}$ with the confidence of the class encoding of the $j$\textsuperscript{th} sample. Following \citet{guideda}, we represent the confidence of the $j$\textsuperscript{th} sample as: 
\begin{equation}
\gamma_j = \exp\left(-\frac{H(\bm{q_j})}{\log_2 C}\right)
\end{equation}

However, we found empirically that doing this in the initial part of the training does not lead to improvements probably because the confidence values the less reliable during the early stages of training so we use this term in conjunction with the decay term $\alpha$. Therefore, the actual confidence score of the $j$\textsuperscript{th} sample is:
\begin{equation}
\gamma_j^{'} = \alpha + (1 - \alpha) \cdot \gamma_j  
\end{equation}

\textbf{Scaling for Class Imbalance} Since the classes may be imbalanced, the class encoding favouring the classes with more representation can dictate the alignment process. For example, if there are many samples in the car class but very few samples in the bus class, then bus samples will experience more frequent attractions towards car samples due to certain semantic similarities between the two compared to other bus samples which can cause bus samples to be misclassified as cars. To counter this, we scale the intensity of alignment towards samples on a batch-wise basis. For each mini batch, we make an estimate of the number of samples belonging to each class based on the $argmax$ of the class encoding and then scale the mask $M_{ij}$ with the class frequency-based scaling factor of the $j$\textsuperscript{th} sample which is given by:

\begin{equation}
w_j = \frac {1}{\alpha + (1 - \alpha) \cdot C_{\text{argmax}(\bm{q_j})} \cdot \frac{C}{B}}
\end{equation}

Where $C_k$ is the number of samples in the current batch estimated to belong to class $k$. $C$ is the total number of classes and $B$ is the batch size.
This means that samples belonging to smaller classes will on average, impart a stronger alignment force, and this effect will come into play later in the training process due to the introduction of $\alpha$

By putting together all of the above terms, the final loss function is given by:

\begin{equation}
L = \sum_{\substack{0 \leq i < bs \\ 0 \leq j < bs \\ i \neq j}} \bm{p_i} \bm{q_j} (M_{ij1} + M_{ij2}) \cdot \gamma_j^{'} \cdot w_j
\end{equation}

Our algorithm is illustrated in Algorithm 1.

\vspace{-10pt}
\begin{figure}
    \centering
    \includegraphics[scale=0.3]{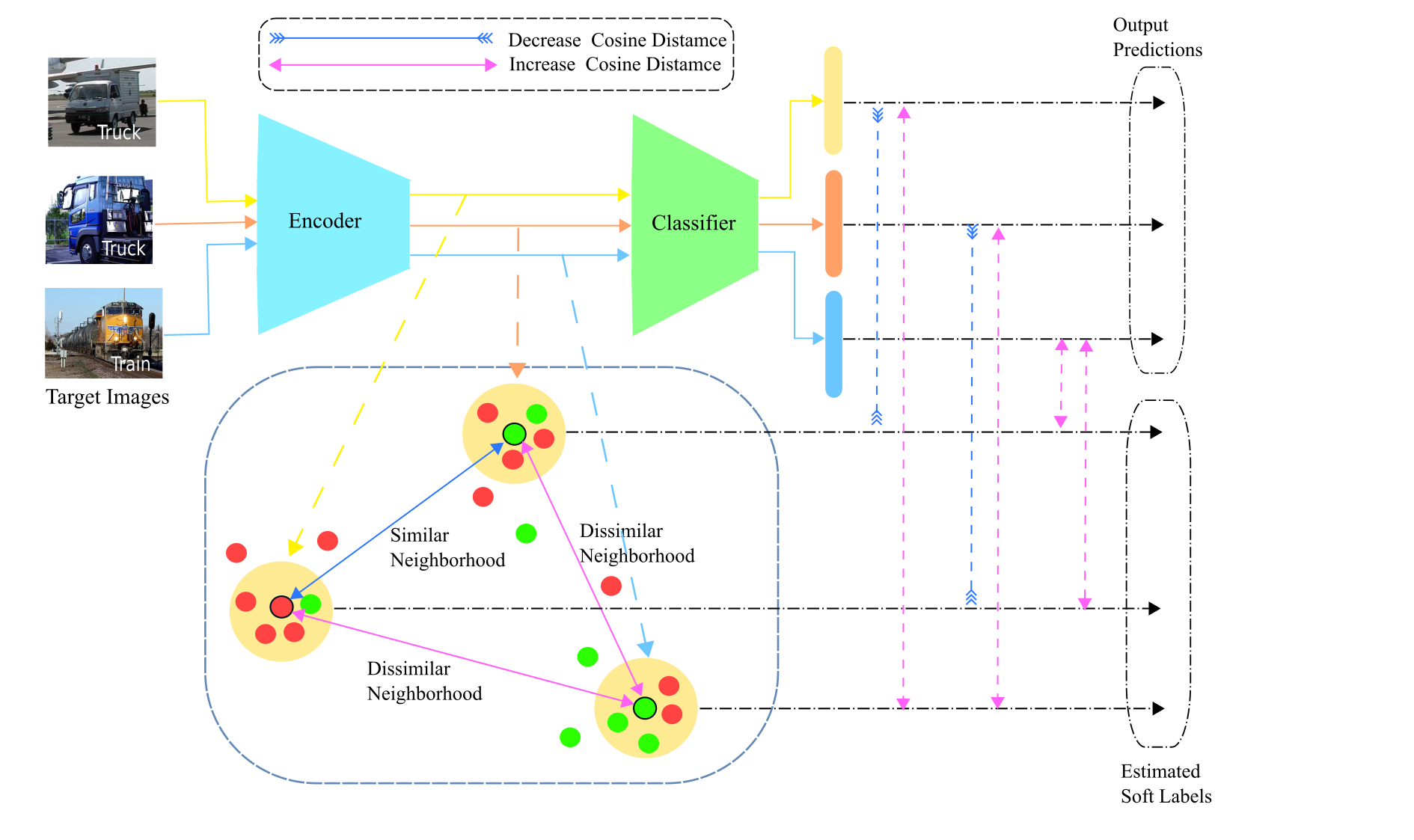}
    \caption{An overview of the proposed Source-Free Domain Adaptation (SFDA) approach. For a batch of data, the encoder maps the input images to the feature space, where we compute the neighbourhood signatures of the samples. Using these neighbourhood signatures, we estimate the class encoding of the samples, and then we optimize the cosine similarity between the output prediction of each sample and the class encoding of all other samples in the batch as per our proposed loss function.}
    \label{fig:example}
    \vspace{-5pt\vspace{-10pt}}
\end{figure}

\begin{algorithm}
\caption{Neighbourhood Signature Contrastive Adaptation}
\label{alg:sfda}
\begin{algorithmic}[1]
\renewcommand{\algorithmicrequire}{\textbf{Input:}}
\Require Source-pretrained model and target data $D_t$
\State Build memory bank storing all target features and predictions
\While{Adaptation}
    \State Sample batch $T$ from $D_t$ and update memory bank
    \For{each sample $\bm{x_i}$ in $T$}
        \State Retrieve $K$-nearest neighbours ($N_i$) of $\bm{x_i}$ and their predictions from memory bank
        \State Compute the neighbourhood signature $\bm{s_i}$ of $\bm{x_i}$ using Eq. 2
        \State Compute the confidence-adaptive class encoding $\bm{q_i}$ of $\bm{x_i}$ using Eq. 3 
        \State Compute the inertia of estimated label $\gamma_j^{'}$ of $\bm{x_i}$ using Eq. 7
        \State Compute the class frequency based scaling factor $w_i$ of $\bm{x_i}$ using Eq. 8
    \EndFor
    \State Update model by minimizing Eq. 9
\EndWhile
\end{algorithmic}
\end{algorithm}

\vspace{10pt}
\section{Experiments}

\textbf{Datasets}. We conduct experiments on three benchmark datasets for image classification: PACS\citep{pacspap}, Office-31\citep{officeapap} and VisDA-C 2017\citep{visdapap}. \textbf{PACS} contains 4 domains (Art-Painting, Cartoon, Photo and Sketch) and 7 object categories with and 9991 images and a large domain shifts due to different styles. \textbf{Office-31} contains 3 domains (Amazon, Webcam, DSLR) with 31 classes and 4,652 images. \textbf{VisDA} is a challenging large scale dataset, having 12 classes with a source domain consisting of 152k synthetic images and a target domain consisting of 55k real object images.

\textbf{Experiment Setup.} To ensure fair comparison with previous methods, we use the same network architecture, training techniques and hyperparameters for all competing methods. For each experiment, the network consists of a feature extractor followed by a classifier. Specifically, we adopt the backbone\citep{resnetpap} of a ResNet-18 for PACS, ResNet-50 for Office-Home and ResNet-101 for VisDA. We adopt SGD with momentum 0.9 and batch size of 64 for all datasets. The learning rate is set to 1e-3 for Office-31 and PACS  and 1e-5 for VisDA. We train for 50 epochs for PACS, 40 epochs for Office-31 and 15 epochs for VisDA.  
There are two hyperparameters, the number of nearest neighbours $K$ and the decay rate $\alpha$. In all our experiments, we set $K$ to 5 and for the decay rate we use: $\alpha = (\frac {1}{2}) ^ \frac {\text{current iter}}{\text{num iter per epoch}}$.

\subsection{Results}

We compare our method with existing SFDA methods on the above-mentioned datasets. For a fair comparison, we reproduce AaD, $\text{SF(DA)}^2$, and SiLAN using their official codes. As shown in Table 1, our method outperforms all others methods on VisDA in terms of average accuracy across the classes. Table 2 shows that our method gives comparable performance to other methods on Office-31 and Table 3 shows that our methods gives significantly better performance compared to other methods on PACS.  

\begin{table}[h]
\centering
\small 
\caption{Comparison of the SFDA methods using ResNet-101 on VisDA2017.}
\vspace{1em}
\begin{tabular}{c|@{\hspace{3pt}}c@{\hspace{3pt}}c@{\hspace{3pt}}c@{\hspace{3pt}}c@{\hspace{3pt}}c@{\hspace{3pt}}c@{\hspace{3pt}}c@{\hspace{3pt}}c@{\hspace{3pt}}c@{\hspace{3pt}}c@{\hspace{3pt}}c@{\hspace{3pt}}c|c}
\hline
\textbf{Method} & \textbf{plane} & \textbf{bcycl} & \textbf{bus} & \textbf{car} & \textbf{horse} & \textbf{knife} & \textbf{mcycl} & \textbf{person} & \textbf{plant} & \textbf{sktbrd} & \textbf{train}& \textbf{truck}  & \textbf{Avg.} \\ \hline \hline 
SHOT\citep{doneeddata} & 94.3 & 88.5 & 80.1 & 57.3 & 93.1 & 94.9 & 80.7 & 80.3 & 91.5 & 89.1 & 86.3 & 58.2 & 82.9 \\ 
HCL\citep{hclpap} & 93.3 & 85.4 & 80.7 & 68.5 & 91.0 & 88.1 & 86.0 & 78.6 & 86.6 & 88.8 & 80.0 & \textbf{74.7} & 83.5 \\
G-SFDA\citep{gsfda} & 96.1 & 88.3 & 85.5 & 74.1 & \textbf{97.1} & 95.4 & 89.5 & 79.4 & 95.4 & 92.9 & 89.1 & 42.6 & 85.4 \\ 
NRC\citep{goodfriend} & 96.8 & 91.3 & 82.4 & 62.4 & 96.2 & 95.9 & 86.1 & 80.6 & 94.8 & 94.1 & 90.4 & 59.7 & 85.9 \\ 
AaD\citep{aada}  & 95.2 & 90.5 & 85.5 & 79.2 & 96.4 & \textbf{96.2} & 88.8 & 80.4 & 93.9 & 91.8 & 91.1 & 55.9 & 87.1 \\ 
DaC\citep{deepadaptclust} & 96.6 & 86.8 & 86.4 & 78.4 & 96.4 & \textbf{96.2} & \textbf{93.6} & \textbf{83.8} & 96.8 & \textbf{95.1} & 89.6 & 50.0 & 87.3 \\ 
$SF(DA)^2$\citep{dalensaug} & 97.2 & 90.9 & 85.0 & 54.0 & 96.7 & 95.9 & 90.1 & 80.9 & 95.0 & 92.7 & 89.2 & 61.4 & 85.8 \\ 
AaD+SiLAN\citep{daoverlook} & 96.4 & 90.4 & 86.1 & 80.0 & 96.9 & 92.3 & 86.8 & 83.0 & 94.7 & 91.7 & 88.5 & 46.1 & 86.1 \\ 

\hline
Ours & \textbf{97.6} & \textbf{91.7} & \textbf{86.7} & \textbf{84.8} & 96.7 & 94.9 & 91.5 & 81.5 & \textbf{97.5} & 93.6 & \textbf{92.5} & 47.0 & \textbf{88.0}\\ 
\hline
\end{tabular}
\end{table}

\begin{table}[h]
\centering
\caption{Comparison of SFDA methods using ResNet-50 on Office-31.}
\vspace{1em}
\begin{tabular}{c|@{\hspace{3pt}}c@{\hspace{3pt}}c@{\hspace{3pt}}c@{\hspace{3pt}}c@{\hspace{3pt}}c@{\hspace{3pt}}c|c}
\hline
\textbf{Method} & \textbf{A$\rightarrow$D} & \textbf{A$\rightarrow$W} & \textbf{D$\rightarrow$W} & \textbf{D$\rightarrow$A} & \textbf{W$\rightarrow$D} & \textbf{W$\rightarrow$A} & \textbf{Avg.} \\ \hline \hline 
SHOT\citep{doneeddata} & 94.0 & 90.1 & 98.4 & 74.7 & 99.9 & 74.3 & 88.6 \\ 
NRC\citep{goodfriend} & 96.0 & 90.8 & \textbf{99.0} & 75.3 & \textbf{100.0} & 75.0 & 89.4 \\
3C-GAN\citep{danosource} & 92.7 & \textbf{93.7} & 98.5 & 75.3 & 99.8 & 77.8 & 89.6 \\ 
HCL\citep{hclpap} & 94.7 & 92.5 & 98.2 & \textbf{75.9} & \textbf{100.0} & 77.7 & \textbf{89.8} \\ 
AaD\citep{aada} & 95.0 & 92.1 & 99.0 & 75.7 & 99.8 & 76.3 & 89.6 \\ 
$SF(DA)^2$\citep{dalensaug} & 94.3 & 88.5 & 80.1 & 57.3 & 93.1 & \textbf{94.9} & 82.9 \\ 
AaD+SiLAN\citep{daoverlook} & 94.3 & 88.5 & 80.1 & 57.3 & 93.1 & \textbf{94.9} & 82.9 \\ 

\hline
Ours & \textbf{96.4} & 93.0 & 98.9 & 75.4 & 99.8 & 73.4 & 89.5 \\ 
\hline
\end{tabular}
\end{table}

\begin{table}[h]
\centering
\caption{Comparison of SFDA methods using ResNet-18 on PACS.}
\vspace{1em}
\begin{tabular}{c|@{\hspace{3pt}}c@{\hspace{3pt}}c@{\hspace{3pt}}c|@{\hspace{3pt}}c@{\hspace{3pt}}c@{\hspace{3pt}}c|c}
\hline
\textbf{Method} & \textbf{P$\rightarrow$A} & \textbf{P$\rightarrow$C} & \textbf{P$\rightarrow$S} & \textbf{A$\rightarrow$P} & \textbf{A$\rightarrow$C} & \textbf{A$\rightarrow$S} & \textbf{Avg.} \\ \hline \hline 
NEL\citep{nelpap} & 82.6 & 80.5 & 32.3 & \textbf{98.4} & 84.3 & 56.1 & 72.4 \\ 
AaD\citep{aada} & 84.8 & 76.0 & 57.2 & 90.0 & 82.4 & 63.2 & 75.6 \\ 
$SF(DA)^2$\citep{dalensaug} & 87.3 & 82.7 & 54.1 & 91.3 & 83.1 & 60.3 & 76.5 \\ 
AaD+SiLAN\citep{daoverlook} & 87.4 & \textbf{85.1} & 43.7 & 90.3 & 83.5 & 64.3 & 75.7 \\ 

\hline
Ours & \textbf{90.9} & 78.2 & \textbf{57.6} & 98.0 & \textbf{90.0} & \textbf{68.6} & \textbf{80.6}\\ 
\hline
\end{tabular}
\end{table}

\subsection{Analysis}

\textbf{Ablation Study} To see the effectiveness of the various components of our method, we conduct ablation studies by removing individual components from the pipeline. The studies are conducted with respect to performance on the PACS dataset. Table 4 shows that each component is crucial in the achieving the best accuracy. 

\vspace{-2em}
\begin{table}[h]
\centering
\caption{Ablation studies of subcomponents of the proposed
method measured by classification accuracy (\%) on PACS.}
\vspace{1em}
\begin{tabular}{l|c} 
\hline
\multicolumn{1}{c|}{\textbf{Method}} &  \textbf{Avg.} \\ \hline \hline 
Ours & 80.6 \\
Ours - Intra-Class Diversity & 80.2 \\
Ours - Inertia of Estimated Labels & 79.7 \\
Ours - Class-Wise Scaling & 78.4 \\
Ours - Confidence-Adaptive Class Encoding & 66.3 \\

\hline
\end{tabular}
\end{table}

\vspace{3em}
\textbf{Effect of number of nearest neighbours} We observe the effect of different choices for the number of nearest neighbours $K$ for performance on the PACS dataset. Figure 2 shows that the performance is robust to various reasonable choices of $K$.

\begin{figure}[h!]
    \centering
\begin{tikzpicture}[scale=0.8]
    \begin{axis}[
        xlabel={\Large Number of Nearest Neighbors (K)},
        ylabel={\Large Accuracy (\%)},
        xmin=2, xmax=7,
        ymin=75, ymax=84,
        xtick={2, 3, 4, 5, 6, 7},
        ytick={75,78,81,84},
        legend pos=south east,
        grid=major,
        width=0.8\linewidth,
        height=0.6\linewidth,
    ]
    \addplot[
        color=blue,
        mark=o,
        mark options={scale=2}
    ] coordinates {
        (2, 79.24)
        (3, 78.92)
        (4, 79.13)
        (5, 80.6)
        (6, 80.25)
        (7, 80.30)
    };
    \legend{Accuracy}
    \end{axis}
\end{tikzpicture}
\caption{Accuracy vs the number of nearest neighbors (K) on PACS}
    \label{fig:accuracy-vs-k}
\end{figure}
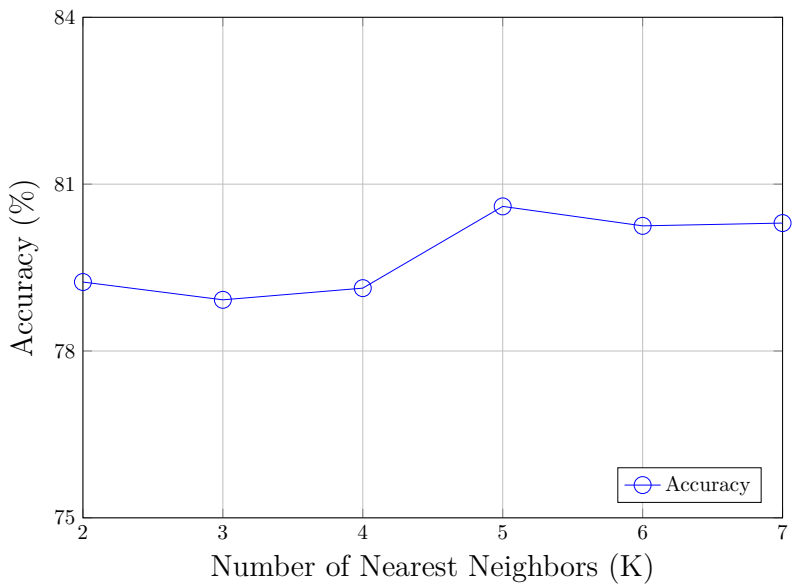

\textbf{Effect of the decay base} In our proposed implementation, we use $\frac{1}{2}$ as the base of the decay factor $\alpha$. We observe the effect of different choices of this base with respect to performance on the PACS dataset. Figure 3 shows that the performance is robust to various reasonable choices for the base of $\alpha$.

\begin{figure}[H]
    \centering
\begin{tikzpicture}[scale=0.8]
    \begin{axis}[
        xlabel={\Large Decay base},
        ylabel={\Large Accuracy (\%)},
        xmin=1/8, xmax=3/4,
        ymin=75, ymax=84,
        xtick={1/8, 2/8, 3/8, 4/8, 5/8, 6/8},
        ytick={75,78,81,84},
        legend pos=south east,
        grid=major,
        width=0.8\linewidth,
        height=0.6\linewidth,
    ]
    \addplot[
        color=red,
        mark=o,
        mark options={scale=2}
    ] coordinates {
        (1/8, 79.7)
        (2/8, 79.0)
        (4/8, 80.6)
        (6/8, 81.3)
    };
    \legend{Accuracy}
    \end{axis}
\end{tikzpicture}
\caption{Accuracy vs the decay base for $\alpha$ on PACS}
    \label{fig:accuracy-vs-base}
\end{figure}
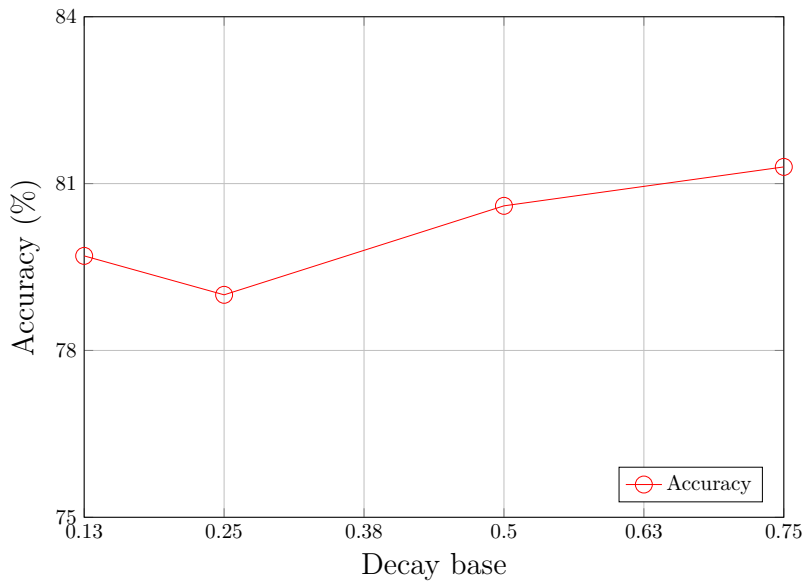

\section{Conclusion}

We proposed to tackle source-free domain adaptation using a novel approach based on aligning predictions of samples based on their neighbourhood signature. We further propose the idea of incorporating intra-class diversity along with prediction inertia and class weight based scaling to design a training objective with a single loss term. Furthermore, we show that the proposed method achieves state-of-the-art performance on several benchmark datasets and also conduct subsequent analysis on the method.

\bibliography{main}

\begin{thebibliography}{38}
\providecommand{\natexlab}[1]{#1}
\providecommand{\url}[1]{\texttt{#1}}
\expandafter\ifx\csname urlstyle\endcsname\relax
  \providecommand{\doi}[1]{doi: #1}\else
  \providecommand{\doi}{doi: \begingroup \urlstyle{rm}\Url}\fi

\bibitem[Ahmed et~al.(2022)Ahmed, Morerio, and Murino]{nelpap}
Waqar Ahmed, Pietro Morerio, and Vittorio Murino.
\newblock Cleaning noisy labels by negative ensemble learning for source-free unsupervised domain adaptation.
\newblock In \emph{2022 IEEE/CVF Winter Conference on Applications of Computer Vision (WACV)}, pp.\  356--365, 2022.
\newblock \doi{10.1109/WACV51458.2022.00043}.

\bibitem[Bousmalis et~al.(2017)Bousmalis, Silberman, Dohan, Erhan, and Krishnan]{pixlevel}
Konstantinos Bousmalis, Nathan Silberman, David Dohan, Dumitru Erhan, and Dilip Krishnan.
\newblock Unsupervised pixel-level domain adaptation with generative adversarial networks.
\newblock In \emph{2017 IEEE Conference on Computer Vision and Pattern Recognition (CVPR)}, pp.\  95--104, 2017.
\newblock \doi{10.1109/CVPR.2017.18}.

\bibitem[Chang et~al.(2017)Chang, Wang, Meng, Xiang, and Pan]{deepadaptclust}
Jianlong Chang, Lingfeng Wang, Gaofeng Meng, Shiming Xiang, and Chunhong Pan.
\newblock Deep adaptive image clustering.
\newblock In \emph{2017 IEEE International Conference on Computer Vision (ICCV)}, pp.\  5880--5888, 2017.
\newblock \doi{10.1109/ICCV.2017.626}.

\bibitem[Chen et~al.(2018)Chen, Kalantidis, Li, Yan, and Feng]{doubleattend}
Yunpeng Chen, Yannis Kalantidis, Jianshu Li, Shuicheng Yan, and Jiashi Feng.
\newblock A2-nets: double attention networks.
\newblock In \emph{Proceedings of the 32nd International Conference on Neural Information Processing Systems}, NIPS'18, pp.\  350–359, Red Hook, NY, USA, 2018. Curran Associates Inc.

\bibitem[Daum{\'e}~III(2007)]{frusda}
Hal Daum{\'e}~III.
\newblock Frustratingly easy domain adaptation.
\newblock In Annie Zaenen and Antal van~den Bosch (eds.), \emph{Proceedings of the 45th Annual Meeting of the Association of Computational Linguistics}, pp.\  256--263, Prague, Czech Republic, June 2007. Association for Computational Linguistics.
\newblock URL \url{https://aclanthology.org/P07-1033/}.

\bibitem[Ganin \& Lempitsky(2015)Ganin and Lempitsky]{unsuback}
Yaroslav Ganin and Victor Lempitsky.
\newblock Unsupervised domain adaptation by backpropagation.
\newblock In \emph{Proceedings of the 32nd International Conference on International Conference on Machine Learning - Volume 37}, ICML'15, pp.\  1180–1189. JMLR.org, 2015.

\bibitem[Ganin et~al.(2016)Ganin, Ustinova, Ajakan, Germain, Larochelle, Laviolette, Marchand, and Lempitsky]{domadver}
Yaroslav Ganin, Evgeniya Ustinova, Hana Ajakan, Pascal Germain, Hugo Larochelle, Fran\c{c}ois Laviolette, Mario Marchand, and Victor Lempitsky.
\newblock Domain-adversarial training of neural networks.
\newblock \emph{J. Mach. Learn. Res.}, 17\penalty0 (1):\penalty0 2096–2030, January 2016.
\newblock ISSN 1532-4435.

\bibitem[Gong et~al.(2012)Gong, Shi, Sha, and Grauman]{geodes}
Boqing Gong, Yuan Shi, Fei Sha, and Kristen Grauman.
\newblock Geodesic flow kernel for unsupervised domain adaptation.
\newblock In \emph{2012 IEEE Conference on Computer Vision and Pattern Recognition}, pp.\  2066--2073, 2012.
\newblock \doi{10.1109/CVPR.2012.6247911}.

\bibitem[Gretton et~al.(2009)Gretton, Smola, Huang, Schmittfull, Borgwardt, and Sch{\"o}lkopf]{coshiftdis}
A.~Gretton, AJ. Smola, J.~Huang, M.~Schmittfull, KM. Borgwardt, and B.~Sch{\"o}lkopf.
\newblock \emph{Covariate shift and local learning by distribution matching}, pp.\  131--160.
\newblock MIT Press, Cambridge, MA, USA, 2009.

\bibitem[He et~al.(2016)He, Zhang, Ren, and Sun]{resnetpap}
Kaiming He, Xiangyu Zhang, Shaoqing Ren, and Jian Sun.
\newblock Deep residual learning for image recognition.
\newblock In \emph{2016 IEEE Conference on Computer Vision and Pattern Recognition (CVPR)}, pp.\  770--778, 2016.
\newblock \doi{10.1109/CVPR.2016.90}.

\bibitem[Huang et~al.(2021)Huang, Guan, Xiao, and Lu]{hclpap}
Jiaxing Huang, Dayan Guan, Aoran Xiao, and Shijian Lu.
\newblock Model adaptation: Historical contrastive learning for unsupervised domain adaptation without source data.
\newblock In A.~Beygelzimer, Y.~Dauphin, P.~Liang, and J.~Wortman Vaughan (eds.), \emph{Advances in Neural Information Processing Systems}, 2021.
\newblock URL \url{https://openreview.net/forum?id=0zXJRJecC_}.

\bibitem[Hwang et~al.(2024)Hwang, Lee, Shin, and Yoon]{dalensaug}
Uiwon Hwang, Jonghyun Lee, Juhyeon Shin, and Sungroh Yoon.
\newblock {SF}({DA})\${\textasciicircum}2\$: Source-free domain adaptation through the lens of data augmentation.
\newblock In \emph{The Twelfth International Conference on Learning Representations}, 2024.
\newblock URL \url{https://openreview.net/forum?id=kUCgHbmO11}.

\bibitem[Kundu et~al.(2020{\natexlab{a}})Kundu, Venkat, M~V, and Babu]{unsfda}
Jogendra Kundu, Naveen Venkat, Rahul M~V, and R.~Babu.
\newblock Universal source-free domain adaptation.
\newblock pp.\  4543--4552, 06 2020{\natexlab{a}}.
\newblock \doi{10.1109/CVPR42600.2020.00460}.

\bibitem[Kundu et~al.(2020{\natexlab{b}})Kundu, Venkat, Ambareesh, RahulM., and Babu]{dainherit}
Jogendra~Nath Kundu, Naveen Venkat, Revanur Ambareesh, V.~RahulM., and R.~Venkatesh Babu.
\newblock Towards inheritable models for open-set domain adaptation.
\newblock \emph{2020 IEEE/CVF Conference on Computer Vision and Pattern Recognition (CVPR)}, pp.\  12373--12382, 2020{\natexlab{b}}.
\newblock URL \url{https://api.semanticscholar.org/CorpusID:215548694}.

\bibitem[Lee et~al.(2019)Lee, Batra, Baig, and Ulbricht]{sliceda}
Chen-Yu Lee, Tanmay Batra, Mohammad Baig, and Daniel Ulbricht.
\newblock Sliced wasserstein discrepancy for unsupervised domain adaptation.
\newblock pp.\  10277--10287, 06 2019.
\newblock \doi{10.1109/CVPR.2019.01053}.

\bibitem[Li et~al.(2017)Li, Yang, Song, and Hospedales]{pacspap}
Da~Li, Yongxin Yang, Yi{-}Zhe Song, and Timothy~M. Hospedales.
\newblock Deeper, broader and artier domain generalization.
\newblock In \emph{{IEEE} International Conference on Computer Vision, {ICCV} 2017, Venice, Italy, October 22-29, 2017}, pp.\  5543--5551. {IEEE} Computer Society, 2017.
\newblock \doi{10.1109/ICCV.2017.591}.
\newblock URL \url{https://doi.org/10.1109/ICCV.2017.591}.

\bibitem[Li et~al.(2020{\natexlab{a}})Li, Jiao, Cao, Wong, and Wu]{danosource}
Rui Li, Qianfen Jiao, Wenming Cao, Hau-San Wong, and Si~Wu.
\newblock Model adaptation: Unsupervised domain adaptation without source data.
\newblock In \emph{2020 IEEE/CVF Conference on Computer Vision and Pattern Recognition (CVPR)}, pp.\  9638--9647, 2020{\natexlab{a}}.
\newblock \doi{10.1109/CVPR42600.2020.00966}.

\bibitem[Li et~al.(2020{\natexlab{b}})Li, Hu, Liu, Peng, Zhou, and Peng]{contraclust}
Yunfan Li, Peng Hu, Zitao Liu, Dezhong Peng, Joey~Tianyi Zhou, and Xi~Peng.
\newblock Contrastive clustering, 2020{\natexlab{b}}.
\newblock URL \url{http://arxiv.org/abs/2009.09687}.
\newblock cite arxiv:2009.09687.

\bibitem[Liang et~al.(2020)Liang, Hu, and Feng]{doneeddata}
Jian Liang, Dapeng Hu, and Jiashi Feng.
\newblock Do we really need to access the source data? {S}ource hypothesis transfer for unsupervised domain adaptation.
\newblock In Hal~Daumé III and Aarti Singh (eds.), \emph{Proceedings of the 37th International Conference on Machine Learning}, volume 119 of \emph{Proceedings of Machine Learning Research}, pp.\  6028--6039. PMLR, 13--18 Jul 2020.
\newblock URL \url{https://proceedings.mlr.press/v119/liang20a.html}.

\bibitem[Litrico et~al.(2023)Litrico, Del~Bue, and Morerio]{guideda}
Mattia Litrico, Alessio Del~Bue, and Pietro Morerio.
\newblock Guiding pseudo-labels with uncertainty estimation for source-free unsupervised domain adaptation.
\newblock pp.\  7640--7650, 06 2023.
\newblock \doi{10.1109/CVPR52729.2023.00738}.

\bibitem[Liu et~al.(2021)Liu, Wang, and Long]{cycleda}
Hong Liu, Jianmin Wang, and Mingsheng Long.
\newblock Cycle self-training for domain adaptation.
\newblock In \emph{Proceedings of the 35th International Conference on Neural Information Processing Systems}, NIPS '21, Red Hook, NY, USA, 2021. Curran Associates Inc.
\newblock ISBN 9781713845393.

\bibitem[Long et~al.(2018)Long, Cao, Wang, and Jordan]{conda}
Mingsheng Long, Zhangjie Cao, Jianmin Wang, and Michael~I. Jordan.
\newblock Conditional adversarial domain adaptation.
\newblock In \emph{Proceedings of the 32nd International Conference on Neural Information Processing Systems}, NIPS'18, pp.\  1647–1657, Red Hook, NY, USA, 2018. Curran Associates Inc.

\bibitem[Lu et~al.(2020)Lu, Yang, Zhu, Liu, Song, and Xiang]{stoiclass}
Zhihe Lu, Yongxin Yang, Xiatian Zhu, Cong Liu, Yi-Zhe Song, and Tao Xiang.
\newblock Stochastic classifiers for unsupervised domain adaptation.
\newblock In \emph{2020 IEEE/CVF Conference on Computer Vision and Pattern Recognition (CVPR)}, pp.\  9108--9117, 2020.
\newblock \doi{10.1109/CVPR42600.2020.00913}.

\bibitem[Peng et~al.(2017)Peng, Usman, Kaushik, Hoffman, Wang, and Saenko]{visdapap}
Xingchao Peng, Ben Usman, Neela Kaushik, Judy Hoffman, Dequan Wang, and Kate Saenko.
\newblock Visda: The visual domain adaptation challenge.
\newblock \emph{ArXiv}, abs/1710.06924, 2017.
\newblock URL \url{https://api.semanticscholar.org/CorpusID:28698351}.

\bibitem[Quionero-Candela et~al.(2009)Quionero-Candela, Sugiyama, Schwaighofer, and Lawrence]{dshiftml}
Joaquin Quionero-Candela, Masashi Sugiyama, Anton Schwaighofer, and Neil~D. Lawrence.
\newblock \emph{Dataset Shift in Machine Learning}.
\newblock The MIT Press, 2009.
\newblock ISBN 0262170051.

\bibitem[Saenko et~al.(2010)Saenko, Kulis, Fritz, and Darrell]{officeapap}
Kate Saenko, Brian Kulis, Mario Fritz, and Trevor Darrell.
\newblock Adapting visual category models to new domains.
\newblock In \emph{Proceedings of the 11th European Conference on Computer Vision: Part IV}, ECCV'10, pp.\  213–226, Berlin, Heidelberg, 2010. Springer-Verlag.
\newblock ISBN 364215560X.

\bibitem[Saito et~al.(2018)Saito, Watanabe, Ushiku, and Harada]{maxdiscrep}
Kuniaki Saito, Kohei Watanabe, Yoshitaka Ushiku, and Tatsuya Harada.
\newblock Maximum classifier discrepancy for unsupervised domain adaptation.
\newblock In \emph{2018 IEEE/CVF Conference on Computer Vision and Pattern Recognition}, pp.\  3723--3732, 2018.
\newblock \doi{10.1109/CVPR.2018.00392}.

\bibitem[Shu et~al.(2018)Shu, Bui, Narui, and Ermon]{dirty}
Rui Shu, Hung Bui, Hirokazu Narui, and Stefano Ermon.
\newblock A {DIRT}-t approach to unsupervised domain adaptation.
\newblock In \emph{International Conference on Learning Representations}, 2018.
\newblock URL \url{https://openreview.net/forum?id=H1q-TM-AW}.

\bibitem[Sun et~al.(2016)Sun, Feng, and Saenko]{coral}
Baochen Sun, Jiashi Feng, and Kate Saenko.
\newblock Return of frustratingly easy domain adaptation.
\newblock In \emph{Proceedings of the Thirtieth AAAI Conference on Artificial Intelligence}, AAAI'16, pp.\  2058–2065. AAAI Press, 2016.

\bibitem[Sun et~al.(2020)Sun, Wang, Liu, Miller, Efros, and Hardt]{testdisshift}
Yu~Sun, Xiaolong Wang, Zhuang Liu, John Miller, Alexei Efros, and Moritz Hardt.
\newblock Test-time training with self-supervision for generalization under distribution shifts.
\newblock In Hal~Daumé III and Aarti Singh (eds.), \emph{Proceedings of the 37th International Conference on Machine Learning}, volume 119 of \emph{Proceedings of Machine Learning Research}, pp.\  9229--9248. PMLR, 13--18 Jul 2020.
\newblock URL \url{https://proceedings.mlr.press/v119/sun20b.html}.

\bibitem[Valiant(1984)]{theolearn}
L.~G. Valiant.
\newblock A theory of the learnable.
\newblock \emph{Commun. ACM}, 27\penalty0 (11):\penalty0 1134–1142, November 1984.
\newblock ISSN 0001-0782.
\newblock \doi{10.1145/1968.1972}.
\newblock URL \url{https://doi.org/10.1145/1968.1972}.

\bibitem[Vapnik(1995)]{stleatheo}
Vladimir~N. Vapnik.
\newblock \emph{The nature of statistical learning theory}.
\newblock Springer-Verlag New York, Inc., 1995.
\newblock ISBN 0-387-94559-8.

\bibitem[Wang et~al.(2024)Wang, Bae, Chen, Zhang, Sigal, and de~Silva]{daoverlook}
Jing Wang, Wonho Bae, Jiahong Chen, Kuangen Zhang, Leonid Sigal, and Clarence~W. de~Silva.
\newblock What has been overlooked in contrastive source-free domain adaptation: Leveraging source-informed latent augmentation within neighborhood context.
\newblock \emph{Transactions on Machine Learning Research}, 2024.
\newblock ISSN 2835-8856.
\newblock URL \url{https://openreview.net/forum?id=iulMde3dP1}.
\newblock Featured Certification.

\bibitem[Wu et~al.(2019)Wu, Long, Wang, Qian, Li, Lin, and Zha]{deepcomprehend}
Jianlong Wu, Keyu Long, Fei Wang, Chen Qian, Cheng Li, Zhouchen Lin, and Hongbin Zha.
\newblock Deep comprehensive correlation mining for image clustering.
\newblock \emph{2019 IEEE/CVF International Conference on Computer Vision (ICCV)}, pp.\  8149--8158, 2019.
\newblock URL \url{https://api.semanticscholar.org/CorpusID:118686508}.

\bibitem[Xie et~al.(2016)Xie, Girshick, and Farhadi]{deepembedanayse}
Junyuan Xie, Ross Girshick, and Ali Farhadi.
\newblock Unsupervised deep embedding for clustering analysis.
\newblock In Maria~Florina Balcan and Kilian~Q. Weinberger (eds.), \emph{Proceedings of The 33rd International Conference on Machine Learning}, volume~48 of \emph{Proceedings of Machine Learning Research}, pp.\  478--487, New York, New York, USA, 20--22 Jun 2016. PMLR.
\newblock URL \url{https://proceedings.mlr.press/v48/xieb16.html}.

\bibitem[Yang et~al.(2021{\natexlab{a}})Yang, Wang, van~de Weijer, Herranz, and Jui]{goodfriend}
Shiqi Yang, Yaxing Wang, Joost van~de Weijer, Luis Herranz, and Shangling Jui.
\newblock Exploiting the intrinsic neighborhood structure for source-free domain adaptation.
\newblock NIPS '21, Red Hook, NY, USA, 2021{\natexlab{a}}. Curran Associates Inc.
\newblock ISBN 9781713845393.

\bibitem[Yang et~al.(2021{\natexlab{b}})Yang, Wang, van~de Weijer, Herranz, and Jui]{gsfda}
Shiqi Yang, Yaxing Wang, Joost van~de Weijer, Luis Herranz, and Shangling Jui.
\newblock { Generalized Source-free Domain Adaptation }.
\newblock In \emph{2021 IEEE/CVF International Conference on Computer Vision (ICCV)}, pp.\  8958--8967, Los Alamitos, CA, USA, October 2021{\natexlab{b}}. IEEE Computer Society.
\newblock \doi{10.1109/ICCV48922.2021.00885}.
\newblock URL \url{https://doi.ieeecomputersociety.org/10.1109/ICCV48922.2021.00885}.

\bibitem[Yang et~al.(2022)Yang, Wang, Wang, Jui, and van~de Weijer]{aada}
Shiqi Yang, Yaxing Wang, Kai Wang, Shangling Jui, and Joost van~de Weijer.
\newblock Attracting and dispersing: a simple approach for source-free domain adaptation.
\newblock In \emph{Proceedings of the 36th International Conference on Neural Information Processing Systems}, NIPS '22, Red Hook, NY, USA, 2022. Curran Associates Inc.
\newblock ISBN 9781713871088.

\end{thebibliography}
\bibliographystyle{tmlr}

\end{document}